\documentclass[10pt,twocolumn,letterpaper]{article}

 \usepackage{iccv}              

\definecolor{iccvblue}{rgb}{0.21,0.49,0.74}
\usepackage[pagebackref,breaklinks,colorlinks,allcolors=iccvblue]{hyperref}
\usepackage{multirow}
\def\paperID{14658} 
\def\confName{ICCV}
\def\confYear{2025}

\title{Efficient Encoder-Free Pose Conditioning and Pose Control for Virtual Try-On}

\author{
Qi Li\textsuperscript{1}\thanks{Equal contribution.} 
\quad
Shuwen Qiu\textsuperscript{2 *}\thanks{Work done during internship at Amazon.}
\quad
Julien Han\textsuperscript{1}\quad
Xingzi Xu\textsuperscript{1,3 \textdagger}\quad
Mehmet Saygin Seyfioglu\textsuperscript{1}\quad
Kee Kiat Koo\textsuperscript{1} \vspace{4pt}\\
Karim Bouyarmane\textsuperscript{1} \vspace{8pt}\\
\textsuperscript{1}Amazon \quad
\textsuperscript{2}University of California, Los Angeles (UCLA) \quad
\textsuperscript{3}Duke University
\vspace{8pt}\\
{\tt\small \{qlimz,hameng,xingzixu,mseyfiog,kiatkoo,bouykari\}@amazon.com}\\
{\tt\small xingzi.xu@duke.edu} \\
{\tt\small jantqiu@cs.ucla.edu} \\
{\color{red}{\tt\small \url{https://pose-vton.github.io/vto-pose-conditioning/}}} \\
{\small \textbf{Work submitted in November 2024 to CVPR 2025}}
}

\begin{document}
\maketitle
\begin{abstract}
As online shopping continues to grow, the demand for Virtual Try-On (VTON) technology has surged, allowing customers to visualize products on themselves by overlaying product images onto their own photos. An essential yet challenging condition for effective VTON is pose control, which ensures accurate alignment of products with the user’s body while supporting diverse orientations for a more immersive experience. However, incorporating pose conditions into VTON models presents several challenges, including selecting the optimal pose representation, integrating poses without additional parameters, and balancing pose preservation with flexible pose control.

In this work, we build upon a baseline VTON model that concatenates the reference image condition without external encoder, control network, or complex attention layers. We investigate methods to incorporate pose control into this pure-concatenation paradigm by spatially concatenating pose data, comparing performance using pose maps and skeletons, without adding any additional parameters or module to the baseline model. Our experiments reveal that pose stitching with pose maps yields the best results, enhancing both pose preservation and output realism. Additionally, we introduce a mixed-mask training strategy using fine-grained and bounding box masks, allowing the model to support flexible product integration across varied poses and conditions.

Our contributions are threefold: 1) We explore different configurations for integrating pose representations into VTON models, 2) We propose a lightweight, parameter-efficient approach for adding pose control, and 3) We enable flexible pose generation through mixed-mask training. Evaluations on public benchmarks demonstrate that our method improves pose preservation and outperforms state-of-the-art models with more complex conditioning frameworks, advancing VTON’s adaptability and realism for diverse real-world applications.
\end{abstract}    
\section{Introduction}
\label{sec:intro}

\begin{figure*}[t]
 \centering
 \includegraphics[width=0.9\textwidth]{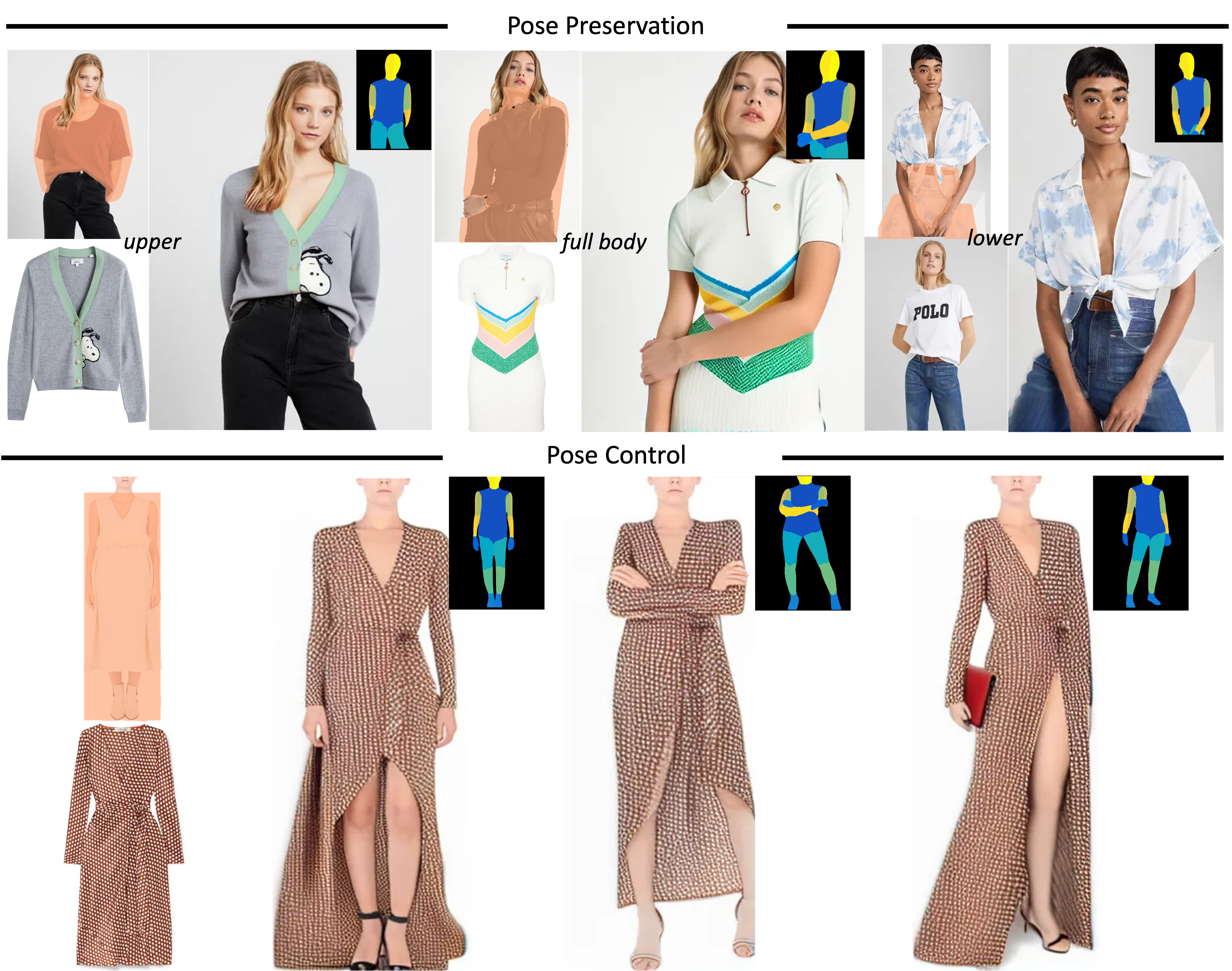}
 \caption{\textbf{Pose preservation and control}. In the first row, pose preservation requires the model to generate natural try-on results while maintaining the user's original pose, even when parts of the body are obscured by a mask. In the second row, pose control tasks the model with generating images of the user in different poses, guided by an input pose image. All examples shown are generated by our model.}
 \label{fig:intro}
\end{figure*}

As online shopping grows in popularity, so does the demand for customers to ``try on'' products virtually by superimposing them onto their own photos. This process, known as Virtual Try-On (VTON) ~\cite{qi2024ditvton, han2024instructvton, xu2025deft}, involves using a source image provided by the user, a specified mask area (of the source image), and an image of the desired product to try on (reference image). The model then seamlessly integrates the product into the masked area of the user provided source image, as shown in~\cref{fig:intro}. Recent advances in UNet-based Latent Diffusion Models (LDMs) ~\cite{rombach2022high, esser2024scaling, nichol2021glide, saharia2022photorealistic} have enabled VTON to be formulated as an image-conditioned inpainting problem, which is fine-tuned with diffusion-based pre-trained weights and has proven to achieve promising generation quality. 

\begin{figure}[t]
  \centering
  \includegraphics[scale=0.35]{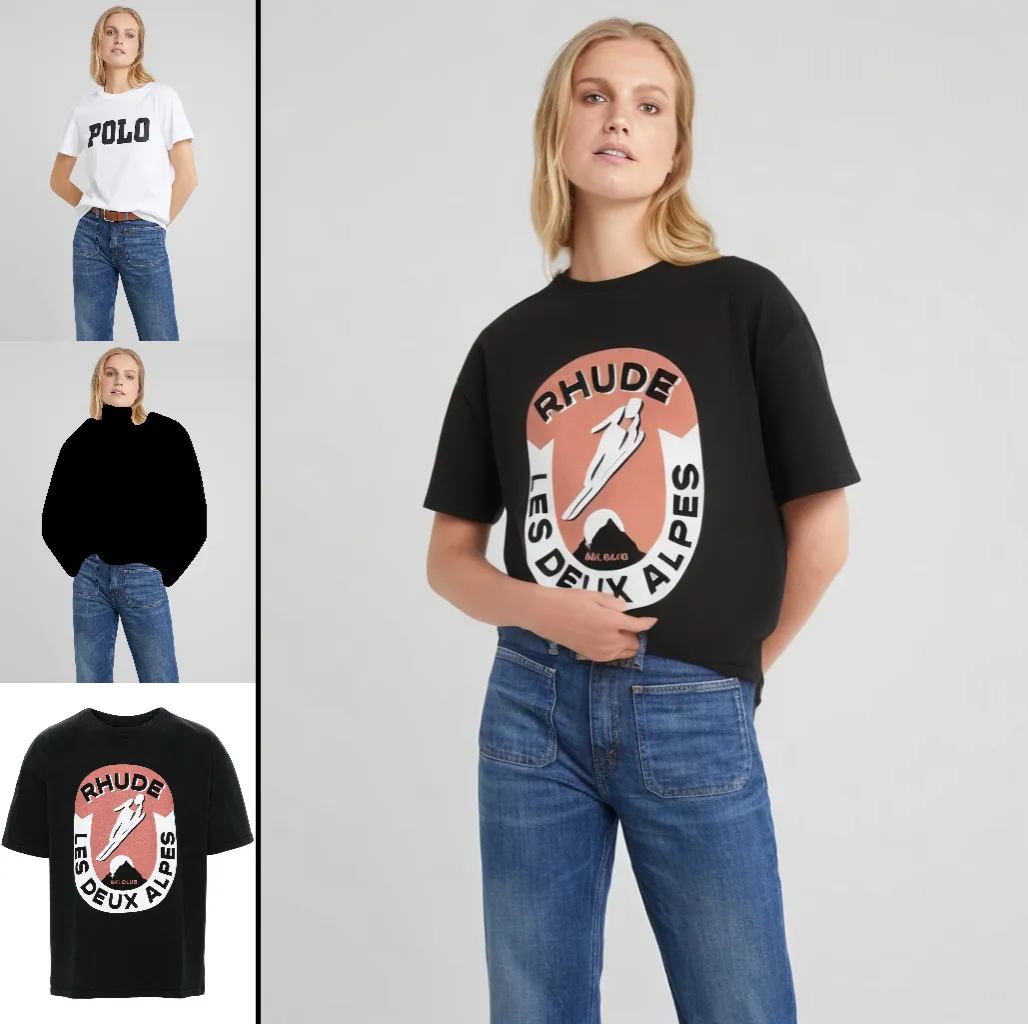}
  \caption{\textbf{A hard case for pose preservation}: When body parts fall within the masked area, the model must hallucinate the pose, and cause natural hand appearance.}
  \label{fig:pose_hard_cases}
\end{figure}

A crucial aspect of VTON tasks is pose control. Pose information is essential for VTON for several reasons. First, when the mask obscures parts of the user's pose in the source image, the model may ``hallucinate'' limbs or other body parts, which is particularly challenging for generative models to render realistically—especially with hands
. This often results in less satisfactory or realistic outputs, as seen in~\cref{fig:pose_hard_cases}. Additionally, incorporating pose control enables VTON models to produce results with various user poses and orientations, enhancing the diversity and realism of user experiences. Although crucial, integrating poses into VTON models presents several challenges: 1) \textbf{Pose Representation}: Common pose representations include pose maps (e.g.~\cite{guler2018densepose}) and skeletons (e.g. ~\cite{8765346}). Current VTON approaches may use either representation or combine them, but it remains unclear which works best for VTON tasks. 2) \textbf{Integration with Model Inputs}: Current models add pose conditions as additional input channels, which requires modifications to the original model weights and additional training parameters. This raises the question of whether pose conditions can be incorporated without extra parameters. 3) \textbf{Pose Control vs. Pose Preservation}: In addition to maintaining the user’s original pose, VTON models must also support customization by generating try-on results with varying user poses. Unlike general image editing, VTON models generate within a masked area, which can hinder realistic outputs if the mask is not aligned with the pose.

In this work, we address these challenges by building on efficient VTON models like DiT-VTON and DEFT-VTON~\cite{qi2024ditvton, han2024instructvton, xu2025deft}, a paradigm that simplified to the extreme the VTON task by removing any need for external image encoders, additional encoding networks ("garment networks", "reference networks", "person network" etc), or complex cross-attention layers. DiT-VTON simply concatenates the masked source image and the reference image along the spatial dimension and feeds the concatenated result into the main diffusion network. This approach achieves simpler and more efficient training while maintaining high-quality generation and detail preservation, outperforming complex architecture alternatives. Motivated by the performance, we investigate whether additional conditions like poses can be further concatenated in the spatial dimension without introducing extra parameters. We use the pose conditioning signal as the additional signal, due to the observed limitation of baseline model in this domain (pose hallucination, hand hallucination when hand is not visible in the reference, etc). We evaluate both pose concatenation and pose stitching techniques, as well as their performance with pose maps and skeletons. Our results indicate that pose stitching with pose maps produces the best outcomes.

Furthermore, we train our model using both fine-grained masks and bounding box masks, enabling it to adapt flexibly to diverse mask conditions. The model learns to fit products precisely within fine-grained masks while allowing more versatile generation with bounding box masks, supporting varied pose-controlled outputs.

Our contributions unfold in three dimensions: 1) We systematically study configurations for integrating different pose representations into VTON models; 2) We propose a simple yet efficient method to incorporate pose conditions without additional channels or extra model parameters; and 3) We enable flexible pose generation through a mixed-mask training strategy. Evaluations on public test benchmarks VITON-HD~\cite{choi2021viton} and DressCode~\cite{morelli2022dress} demonstrate that our method enhances pose preservation and surpasses state-of-the-art (SOTA) VTON models that rely on more complex pose-conditioning frameworks.

\section{Related Work}
\label{sec:related work}

\textbf{Image Virtual Try-On with Diffusion Models}
Latent Diffusion Models (LDMs)~\cite{rombach2022high, esser2024scaling, nichol2021glide, saharia2022photorealistic} have demonstrated strong generative performance across tasks such as text-to-image generation, image inpainting, and image editing. To adapt diffusion models for VTON tasks, WarpDiffusion~\cite{li2023warpdiffusion} integrates a warping module~\cite{han2018viton, wang2018toward, xie2023gp, ge2021parser} that aligns the garment with the masked area and uses a pre-trained text-to-image diffusion model enhanced by a novel information and local garment feature attention mechanism. TryOnDiffusion~\cite{zhu2023tryondiffusion} introduces a dual-UNet structure with cross-attention to implicitly incorporate the warping process, linking the garment condition with the model. For better garment representation infusion into the denoising UNet, LaDI-VTON~\cite{morelli2023ladi} applies a textual inversion module to embed garment-specific information.

Moving beyond warping, StableVTON~\cite{kim2024stableviton} incorporates a supervision signal that aligns the attention map with the warped clothing, while IDM-VTON~\cite{choi2024improving} combines textual inversion and a dual-UNet structure to achieve robust try-on results with diverse, real-world images. Further extensions in VTON, such as Anyfit~\cite{li2024anyfit}, MV-VTON~\cite{wang2024mv}, and Wear-Any-Way~\cite{chen2024wear}, introduce capabilities for multi-garment, multi-view, and interactive editing. Meanwhile, some works shift towards a single UNet structure. CatVTON~\cite{chong2024catvton} reduces complexity by removing extra garment modules and cross-attention layers; TPD~\cite{yang2024texture} uses a single UNet with an additional mask channel to refine the mask input; MMTryon~\cite{zhang2024mmtryon} leverages additional encoders for multi-modal, multi-reference generation, while M\&M VTO~\cite{zhu2024m} incorporates a Diffusion Transformer and supplementary encoders to enable style-controlled generation.

\textbf{Adding Control to Diffusion Models}
Diffusion models for text based image editing have shown superior performances. However, abstract text prompting cannot provide details about visual concepts such as poses, gestures, textures, etc. A conventional way to introduce control signals into the diffusion models involves using a copy of the UNet encoder as the ControlNet~\cite{zhang2023adding}, which encodes the control images such as sketches, poses, and depth information into the denosing UNet. Text inversion with IP Adapter~\cite{ye2023ip} and DreamBooth~\cite{ruiz2023dreambooth} offers another alternative for additional conditioning. In VTON task specifically, pose signal is an important factor to help preserve and generate various poses for the given subject. Most works add additional channels to accommodate the pose images~\cite{kim2024stableviton, choi2024improving, chen2024zero, morelli2023ladi, yang2024texture, wang2024mv}. Meanwhile, two types of pose representations — pose maps (e.g.~\cite{guler2018densepose}) and skeletons (e.g.~\cite{8765346}) — are mixedly used without validations. In this work, we take a closer look at the pose conditioning and study the performance of different pose representations and the effective and efficient way to incorporate pose signals into the generation process.

\section{Method}
\begin{figure*}[t]
  \centering
  \includegraphics[width=0.8\textwidth]{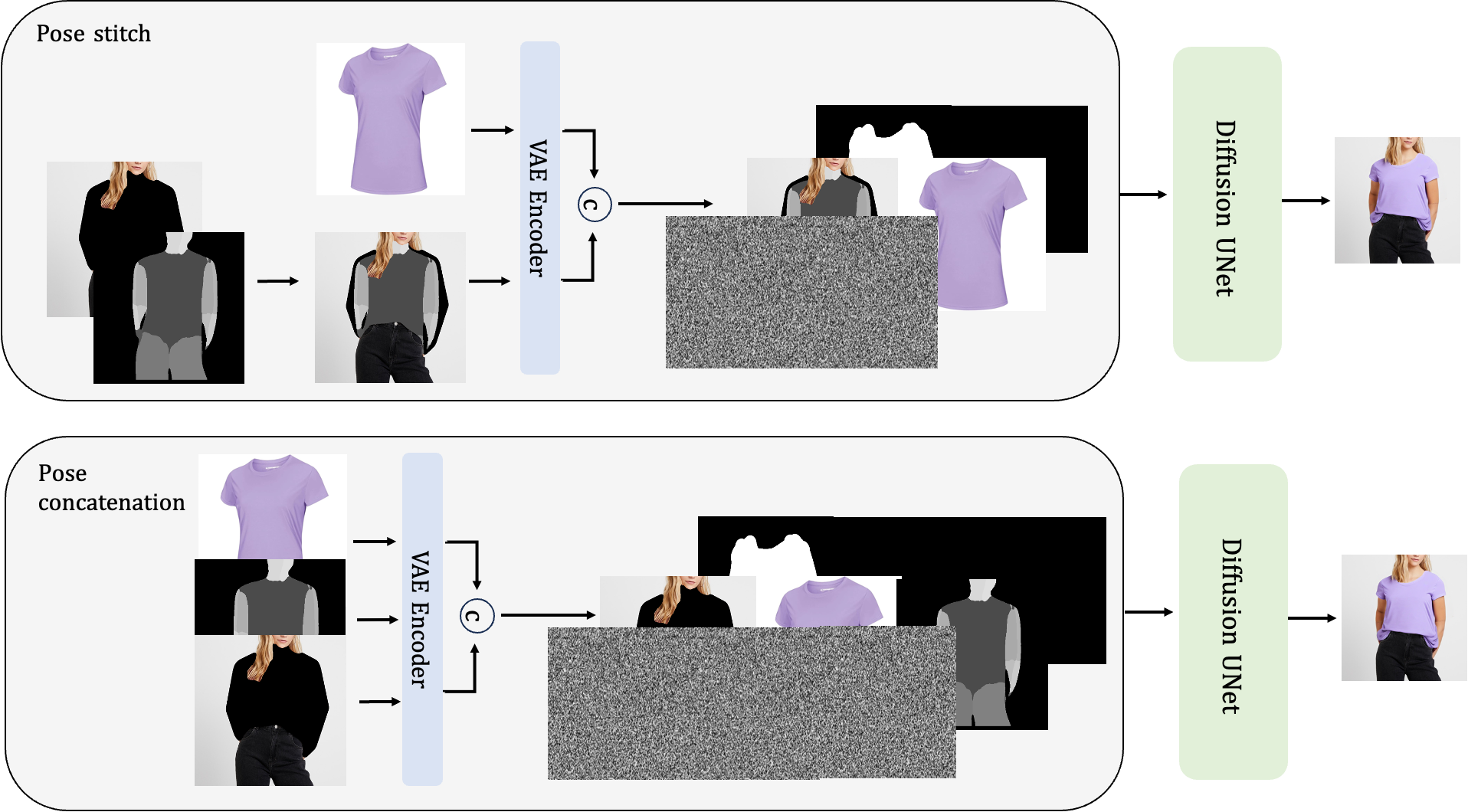}
  \caption{\textbf{Adding pose conditions}. The top image illustrates the construction of the input using pose stitching, while the bottom image shows the model input with pose concatenation.}
  \label{fig:pose_survey}
\end{figure*}

\subsection{Preliminary}
\subsubsection{Latent Diffusion Models}  
Latent Diffusion Models \cite{rombach2022high,podell2023sdxl} operate by learning a gradual denoising process through a series of timesteps that reverse a noise-injection process. Given an image \( x_0 \), the forward diffusion process adds Gaussian noise step-by-step, creating a latent variable \( x_t \) at each timestep \( t \). This process is represented as:
\[
x_t = \sqrt{\alpha_t} \, x_0 + \sqrt{1 - \alpha_t} \, \epsilon,
\]
where \( \alpha_t \) is a noise schedule parameter and \( \epsilon \sim \mathcal{N}(0, I) \) is Gaussian noise. The model is then trained to predict the noise \( \epsilon \) and gradually denoise \( x_t \) back to \( x_0 \) over multiple steps. Stable Diffusion optimizes this process by working within a compressed latent space, enabling high-resolution generation with reduced computational requirements.

\subsubsection{Denoising Diffusion Implicit Models (DDIM)}  
DDIM \cite{song2020denoising} builds on diffusion models by introducing a deterministic sampling method that accelerates the reverse diffusion process. Unlike traditional diffusion, which follows a Markovian sequence, DDIM allows for non-Markovian sampling with fewer steps. The reverse process in DDIM can be written as:
\[
x_{t-1} = \sqrt{\alpha_{t-1}} \, \hat{x}_0 + \sqrt{1 - \alpha_{t-1}} \, \epsilon_t,
\]
where \( \hat{x}_0 \) is an estimate of the original image at each timestep. This approach enables DDIM to achieve high-quality results with significantly fewer sampling steps, enhancing efficiency while maintaining image fidelity.

\subsection{Adding Pose Conditions into the Diffusion Models}
Virtual try-on generates a composite image of a person wearing a specified garment. The input includes a person image \( I_p \in \mathbb{R}^{3\times H \times W} \), a garment image \( I_g \in \mathbb{R}^{3\times H \times W} \), and a binary mask \( M \) defining the editable area. The inpainted person image, which serves as input, is denoted as \( I_m = I_p \otimes M \). To enable precise pose control, a pose image \( I_c \) is additionally extracted from \( I_p \). These inputs—\( I_m \), \( I_g \), and \( I_c \)—are encoded into the latent space by a VAE encoder \( E \): $x_m = E(I_m), x_g = E(I_g), x_c = E(I_c)$
where \( x \in \mathbb{R}^{4 \times H/8 \times W/8} \). The mask \( M \) is interpolated to the size \( m \in \mathbb{R}^{H/8 \times W/8} \) to align with the latent representations.

\subsubsection{Stitching Conditions in the Spatial Dimension}
As shown in recent work~\cite{qi2024ditvton, xu202DEFTVTON, chong2024catvton, yang2024texture, choi2024improving}, stitching a reference image with the masked image along the spatial dimension can effectively preserve text and logo details in garments. Specifically, Stable Diffusion (SD) inpainting models extend the original 4-channel noisy image into 9 channels by concatenating \( z_t \), \( x_m \), and \( m \) in the channel dimension, denoted as \( \text{concat}_{c} \). To integrate an additional garment image \( x_g \), it is concatenated alongside the masked image in the spatial dimension, denoted as \( \text{concat}_{s} \). Additional channels are padded with black images \( x_\alpha \in \mathbb{R}^{4 \times H/8 \times W/8} \) and \( x_\beta \in \mathbb{R}^{H/8 \times W/8} \): $x_{\text{masked}}= \text{concat}_{s}([x_m, x_g]), x_{\text{mask}}= \text{concat}_{s}([m, x_\beta]), x_{\text{noisy}}= \text{concat}_{s}([x_t, x_\alpha])$.

The final input to the transformer blocks is 
$x_{\text{in}} = \text{concat}_{c}[x_{\text{noisy}}, x_{\text{masked}}, x_{\text{mask}}]$.
The model's output is then cut in half.

\begin{table*}[t]
\centering
\caption{Quantitative comparison on VTION-HD dataset of different methods for combining pose representation (joints vs. pose maps), integration methods (stitching vs. concatenation), and color modes (color vs. grayscale). }\label{tab:res pose}
\begin{tabular}{lrrrrlrrrr}
\toprule
\multirow{2}{*}{Pose conditions} & \multicolumn{4}{c}{Stable Diffusion v1.5} & \multirow{7}{*}{} & \multicolumn{4}{c}{Stable Diffusion XL} \\ \cmidrule(lr){2-5} \cmidrule(lr){6-10} 
 & \multicolumn{1}{c}{SSIM$\uparrow$} & \multicolumn{1}{c}{LPIPS$\downarrow$} & \multicolumn{1}{c}{FID$\downarrow$} & \multicolumn{1}{c}{KID$\downarrow$} &  & \multicolumn{1}{c}{SSIM$\uparrow$} & \multicolumn{1}{c}{LPIPS$\downarrow$} & \multicolumn{1}{c}{FID$\downarrow$} & \multicolumn{1}{c}{KID$\downarrow$} \\ \midrule 
Ours w/ Joints Stitch & 0.8861 & \underline{0.0958} & \underline{9.052} & \underline{1.175} &  & \underline{0.8864} & 0.0941 & \underline{10.091} & \underline{2.088} \\
Ours w/ Joints Concat & 0.8781 & 0.1087 & 10.467 & 1.725 &  & 0.8695 & 0.1078 & 12.517 & 3.734 \\ \midrule
Ours w/ Pose Concat & 0.8870 & 0.1148 & 12.623 & 3.277 &  & 0.8861 & 0.0928 & 11.895 & 3.144 \\
Ours w/ Pose Concat Gray & \underline{0.8879} & 0.1120 & 12.449 & 3.256 &  & 0.8856 & \underline{0.0918} & 11.809 & 2.975 \\
Ours w/ Pose Stitch Gray & \textbf{0.9053} & \textbf{0.0694} & \textbf{8.646} & \textbf{0.872} &  & \textbf{0.8993} & \textbf{0.0766} & \textbf{9.341} & \textbf{1.367} \\ \bottomrule
\end{tabular}
\end{table*}

\begin{table*}[t]
\centering
\caption{Quantitative comparison with baselines on the VITON-HD and DressCode test sets. Models labeled with /w SD and SDXL use Stable Diffusion v1.5 and Stable Diffusion XL as the backbone, respectively.}\label{tab:res baseline}
\begin{tabular}{lccccccccc}
\toprule
\multicolumn{1}{c}{\multirow{2}{*}{Models}} & \multicolumn{4}{c}{VITONHD} & \multicolumn{1}{l}{} & \multicolumn{4}{c}{DressCode} \\ \cmidrule(lr){2-5} \cmidrule(lr){6-10} 
\multicolumn{1}{c}{} & SSIM$\uparrow$ & LPIPS $\downarrow$& FID$\downarrow$ & KID$\downarrow$ &  & SSIM$\uparrow$ & LPIPS $\downarrow$& FID $\downarrow$& KID $\downarrow$\\ \midrule
StableVTON~\cite{kim2024stableviton} & 0.8543 & 0.0905 & 11.054 & 3.914 &  & - & - & - & - \\
LaDI-VTON~\cite{morelli2023ladi} & 0.8603 & 0.0733 & 14.648 & 8.754 &  & 0.7656 & 0.2366 & 10.676 & 5.787 \\
IDM-VTON~\cite{choi2024improving} & 0.8499 & \underline{0.0603} & 9.842 & 1.123 &  & 0.8797 & 0.0563 & 9.546 & 4.320 \\
CatVTON~\cite{chong2024catvton} & 0.8704 & \textbf{0.0565} & \underline{9.015} & \underline{1.091} &  & 0.8922 & \textbf{0.0455} & 6.137 & 1.403 \\ \midrule
Ours w/ SD & \textbf{0.9053} & 0.0694 & \textbf{8.646} & \textbf{0.872} &  & \textbf{0.9277} & \underline{0.0510} & \textbf{5.103} & \textbf{0.951} \\
Ours w/ SDXL & \underline{0.8993} & 0.0766 & 9.341 & 1.367&  & \underline{0.9252} & 0.0532 & \underline{5.545} & \underline{1.358} \\ \bottomrule 
\end{tabular}
\end{table*}
\subsection{Adding Pose Constraints}

Above, we considered the minimal conditions for the VTON task. Additional conditions, such as pose and depth information, can help better preserve the pose within the masked area. In practice, there are two primary representations of pose: pose maps and skeletons. A pose map is a dense representation, often visualized as a heatmap or color-coded overlay, where each pixel intensity indicates the likelihood of a specific body part being located at that position (see, e.g.~\cite{guler2018densepose}). Skeleton joints, on the other hand, provide a sparse representation of body pose, typically using a set of key points (or "joints") representing crucial body parts such as elbows, knees, wrists, and shoulders. These joints are connected by lines to form a skeleton-like structure, representing the person's pose (see, e.g.~\cite{8765346}). For each type of representation, we explore two possible methods in~\cref{fig:pose_survey} without introducing additional training parameters to the model, making them adaptable to any VTON model:
\begin{itemize}
    \item \textbf{Concatenate Pose in the Spatial Dimension.} A straightforward approach to incorporating additional conditions is to concatenate them after the reference image: 
    $ x_{\text{masked}}=\text{concat}_{s}([x_m, x_g, x_c]), x_{\text{mask}}=\text{concat}([m, x_\beta, x_\beta]),x_{\text{noisy}} = \text{concat}([x_t, x_\alpha, x_\alpha])$

    The final output only utilizes the first third along the spatial dimension.
    
    \item \textbf{Stitch Pose into the Masked Area.} Since the pose constraint originates from the source image, we can integrate the pose and masked image as a single image using $I_m = I_p \otimes M + I_c \otimes (1 - M)$.
\end{itemize}

\subsubsection{Masking Strategy}
In inpainting tasks, the mask plays a crucial role in marking the boundary of the context and conveying the user's intent. For various pose conditioning scenarios, we consider corresponding masking strategies:

\begin{itemize}
    \item \textbf{Fine-grained Mask:} In this scenario, users employ the mask to precisely define the boundary of the inpainting area, indicating how the clothing should be overlaid onto the person. This strategy is primarily used for pose preservation.

    \item \textbf{Bounding Box Mask:} When users aim to adjust the original pose by providing a pose image not derived from the source image, the fine-grained mask may not effectively capture the intended boundary. To offer greater flexibility in model generation for different poses, a bounding box is obtained by find the minimal rectangular that covers the fine-grained mask. It is used to represent the approximate area being edited. 
\end{itemize}

During training, the fine-grained and bounding box masks are applied alternately to help the model recognize and differentiate between these distinct intentions.


\begin{figure*}[t]
  \centering
  \includegraphics[width=0.9\textwidth]{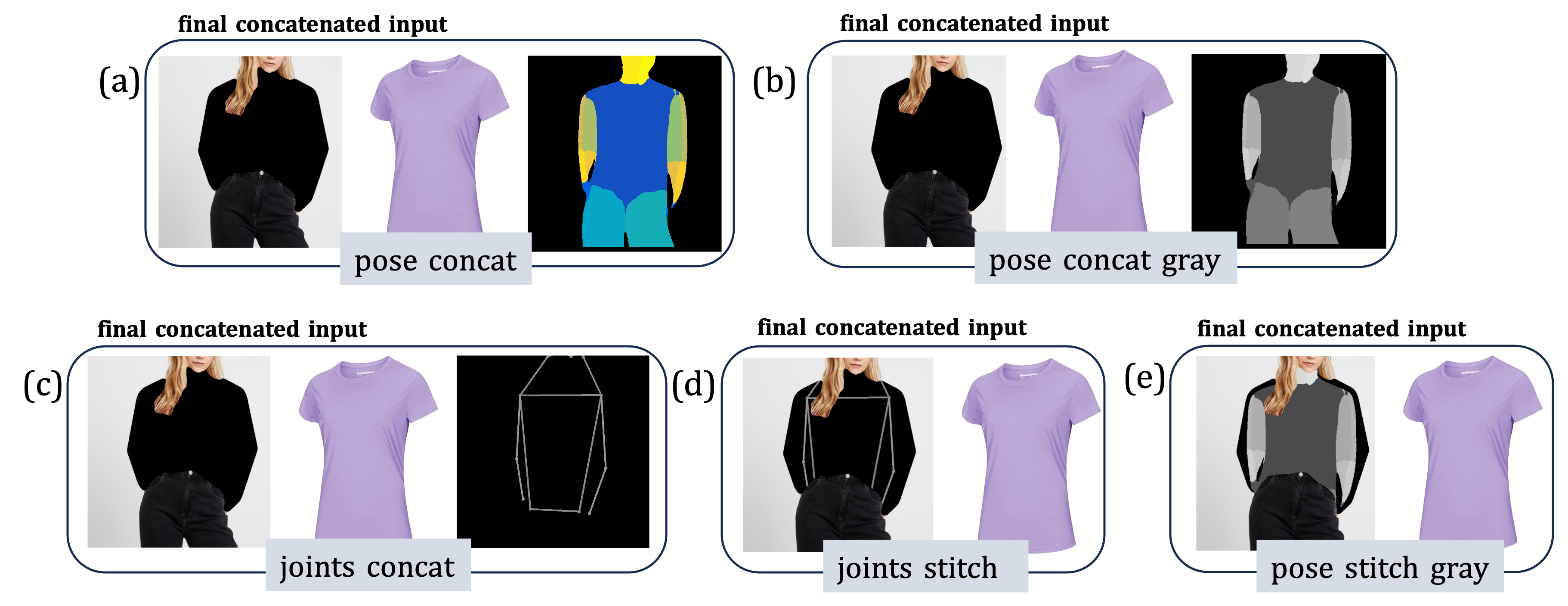}
  \caption{Illustration of different methods for combining pose representation (joints vs. pose map), integration techniques (stitching vs. concatenation), and color modes (color vs. grayscale).}
  \label{fig:five combs}
\end{figure*}
\begin{figure*}[t]
  \centering
  \includegraphics[width=0.9\textwidth]{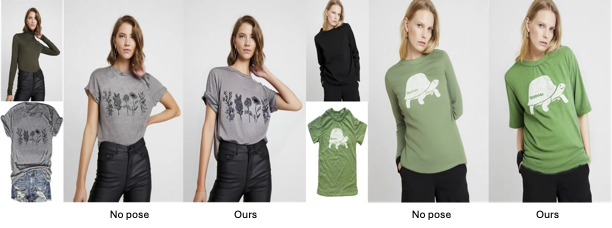}
  \caption{Qualitative comparison of baseline pose-free/pose-less models and our methods for pose preservation.}
  \label{fig:qual pose pre}
\end{figure*}
\section{Experiment}
\subsection{Datasets}

We conduct our tests on two publicly available virtual try-on datasets: VITON-HD~\cite{choi2021viton} and DressCode~\cite{morelli2022dress}. Each input image is padded and resized to a resolution of 512$\times$512. VITON-HD contains 2,032 image pairs of upper-body garments in its test split. We utilize the preprocessed masks provided in this dataset. DressCode comprises 5,400 testing pairs across three garment categories (upper-body, lower-body, and full dresses).

\subsection{Pose Representations}
For pose maps, VITON-HD test set provides preprocessed pose maps directly within the dataset, while for DressCode, the results extracted from a pose map representation model (similar to~~\cite{guler2018densepose}). We then apply the same color map as VITON-HD to generate pose visualization images. For joint skeletons, we generate joint points using a joints pose representation model.

\subsection{Adding Pose Conditions}
To integrate pose conditions into the model inputs, we experiment with two pose representations (skeleton joints and pose maps) and two input formats (concatenation and stitching). Additionally, preliminary studies suggest that color in pose images can interfere with generation quality. Therefore, we test both color and grayscale formats for pose and skeleton images, resulting in five control groups, as illustrated in~\cref{fig:five combs}:

\begin{itemize} 
\item \textbf{Joints Stitch}: Stitching the skeleton joints image into the masked regions. 
\item \textbf{Joints Concat}: Concatenating the skeleton joints image along the spatial dimension, adjacent to the reference image. 
\item \textbf{Pose Concat}: Concatenating the pose map image along the spatial dimension, adjacent to the reference image. 
\item \textbf{Pose Concat Gray}: Concatenating the grayscale-converted pose map image along the spatial dimension, adjacent to the reference image. 
\item \textbf{Pose Stitch Gray}: Stitching the grayscale-converted pose map image into the masked regions. 
\end{itemize}

\begin{figure*}[t]
  \centering
  \includegraphics[width=0.8\textwidth]{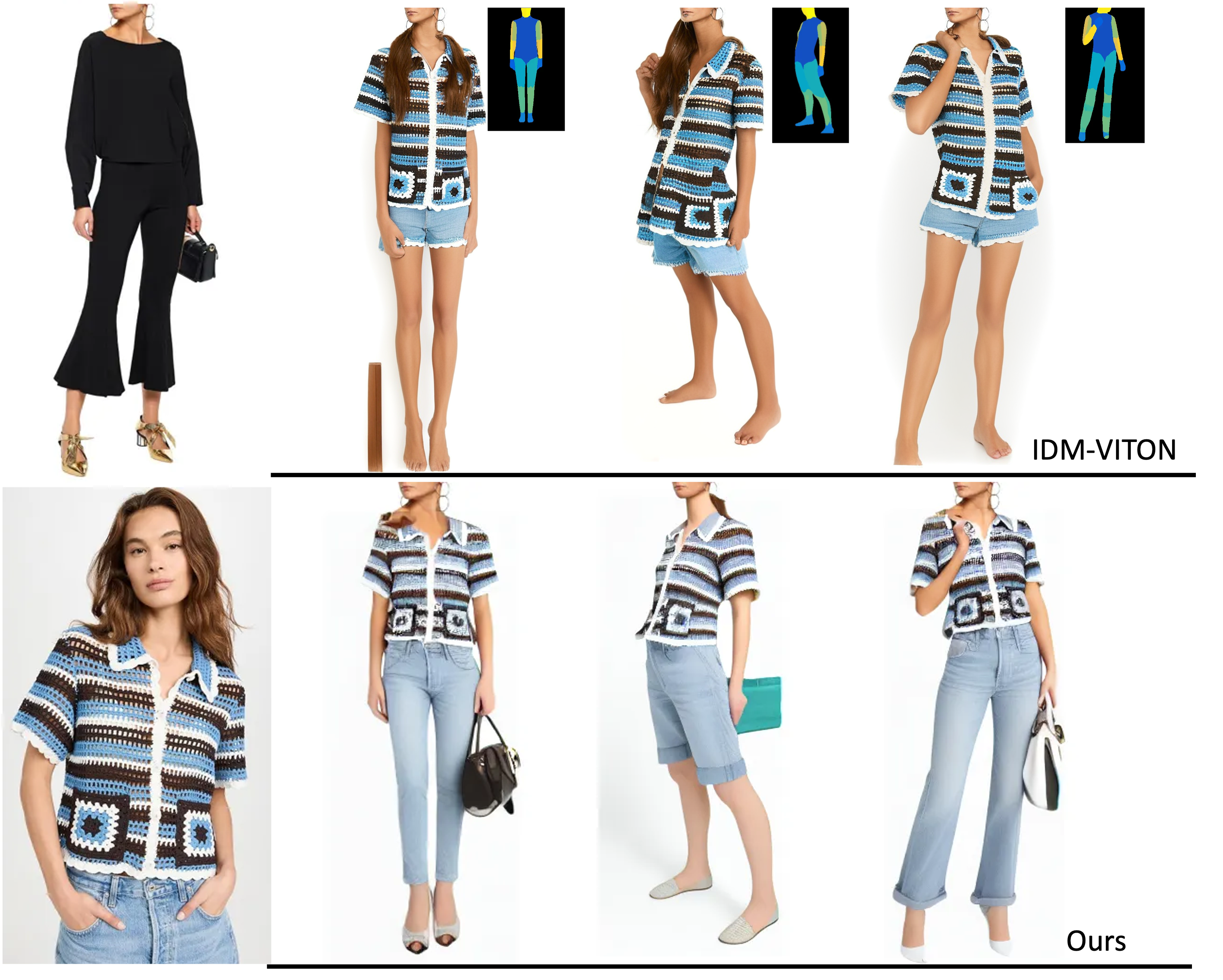}
  \caption{Qualitative comparison of IDM-VTON and our methods for pose-controlled try-on generation.}
  \label{fig:qual pose control}
\end{figure*}
\subsection{Implementation Details}
We employ Stable Diffusion v1.5 
as our diffusion backbone models~\cite{rombach2022high, podell2023sdxl}. The model is trained with a learning rate of 1e-5 and a batch size of 8. We use DDIM with 25 sampling steps and a guidance scale of 5. Both diffusion models are fine-tuned for 20 epochs on 4 NVIDIA H100 80G GPUs. During training, we use 50\% fine-grained masks and 50\% bounding box masks. 

\subsection{Baselines and Evaluation Metrics}
We select the latest baseline results from CatVTON~\cite{chong2024catvton}, StableVTON~\cite{kim2024stableviton}, LaDI-VTON~\cite{morelli2023ladi}, IDM-VTON~\cite{choi2024improving}
To evaluate the model’s output, we employ the widely used evaluation metrics: Structural SIMilarity Index (SSIM)~\cite{wang2004image} and Learned Perceptual Image Patch Similarity (LPIPS)~\cite{zhang2018unreasonable}, comparing generated results to ground truth images. In the unpaired setting, where the garment in the person image differs from the input garment image, we utilize Fréchet Inception Distance (FID)~\cite{zhang2018unreasonable} and Kernel Inception Distance (KID)~\cite{binkowski2018demystifying} as additional metrics. Our implementation follows that in~\cite{morelli2023ladi}.

\section{Results}

\begin{figure*}[t]
  \centering
  \includegraphics[width=0.8\textwidth]{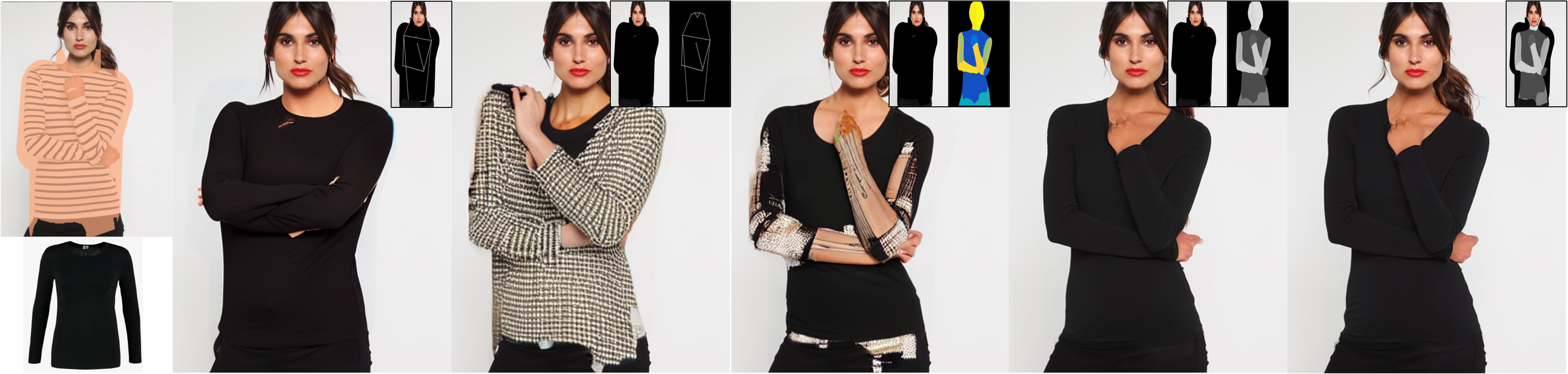}
  \caption{Qualitative comparison of the five methods for pose integration into the model input.}
  \label{fig:ablation}
\end{figure*}

\subsection{Quantitative Results}

\subsubsection{Pose Maps vs Skeleton Joints} Comparing pose concatenation rows with joints and pose maps (see~\cref{tab:res pose}), we observe that pose maps consistently outperform skeleton joints across all metrics. For stitched results, the grayscale pose maps row yield superior outcomes than the joints stitching row. This suggests that pose maps capture richer pose information, which enhances the quality of the generated results.

\subsubsection{Concatenating vs Stitching} Within each pose representation, we compare the concatenation and stitching methods. As shown in~\cref{tab:res pose}, stitching outperforms concatenation for both representations. Additionally, converting pose maps to grayscale does not enhance performance in concatenation methods but significantly improves stitching methods. This finding implies that direct concatenation may not fully leverage the pose information. To better integrate the pose map, stitching with grayscale pose maps is preferred.

\subsubsection{SD vs SDXL} In~\cref{tab:res pose}, a comparison between the left and right sections shows that SDXL and SD v1.5 achieve comparable results. Observations regarding pose representations and integration methods are consistent across both model architectures.

\subsubsection{Comparison with Baseline Results}
In~\cref{tab:res baseline}, we explore the potential benefits of pose conditioning on VTON performance. Building on the original encoder-free VTON architectures in~\cite{qi2024ditvton, xu202DEFTVTON} we stitch the grayscale pose map into the masked area and test both SD v1.5 and SDXL inpainting models as backbones. Integrating the pose image without introducing additional training parameters consistently improves the original results and outperforms other baselines that add an extra channel for the pose image. This comparison demonstrates that stitching grayscale maps can enhance generation results in an effective manner.

\begin{figure*}[h]
  \centering
  \includegraphics[width=0.6\textwidth]{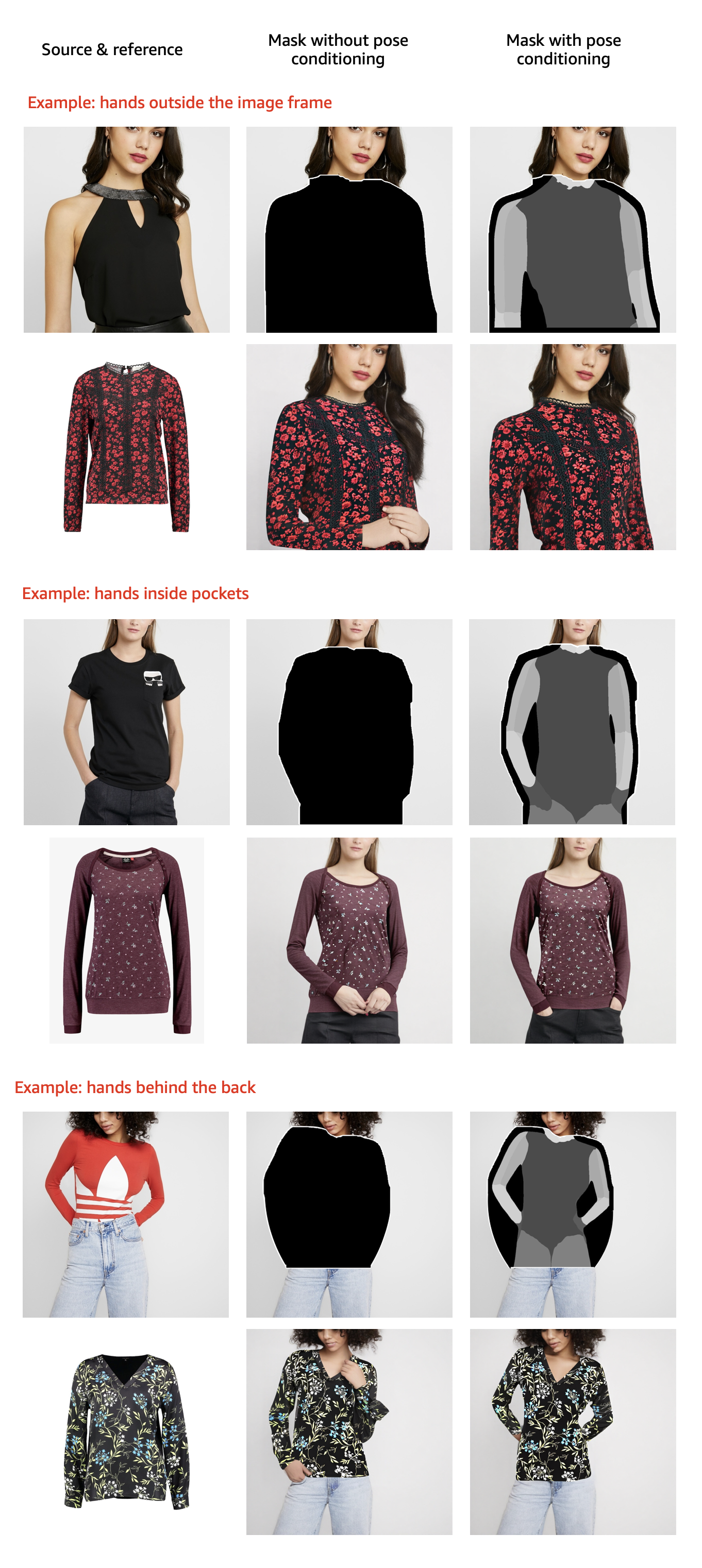}
  \caption{Additional comparisons of baseline pose-free/pose-less models and our methods for pose preservation. We show typical failure cases of pose-free model.}
  \label{fig:posecond}
\end{figure*}

\begin{figure}[h]
  \centering
  \includegraphics[scale=0.5]{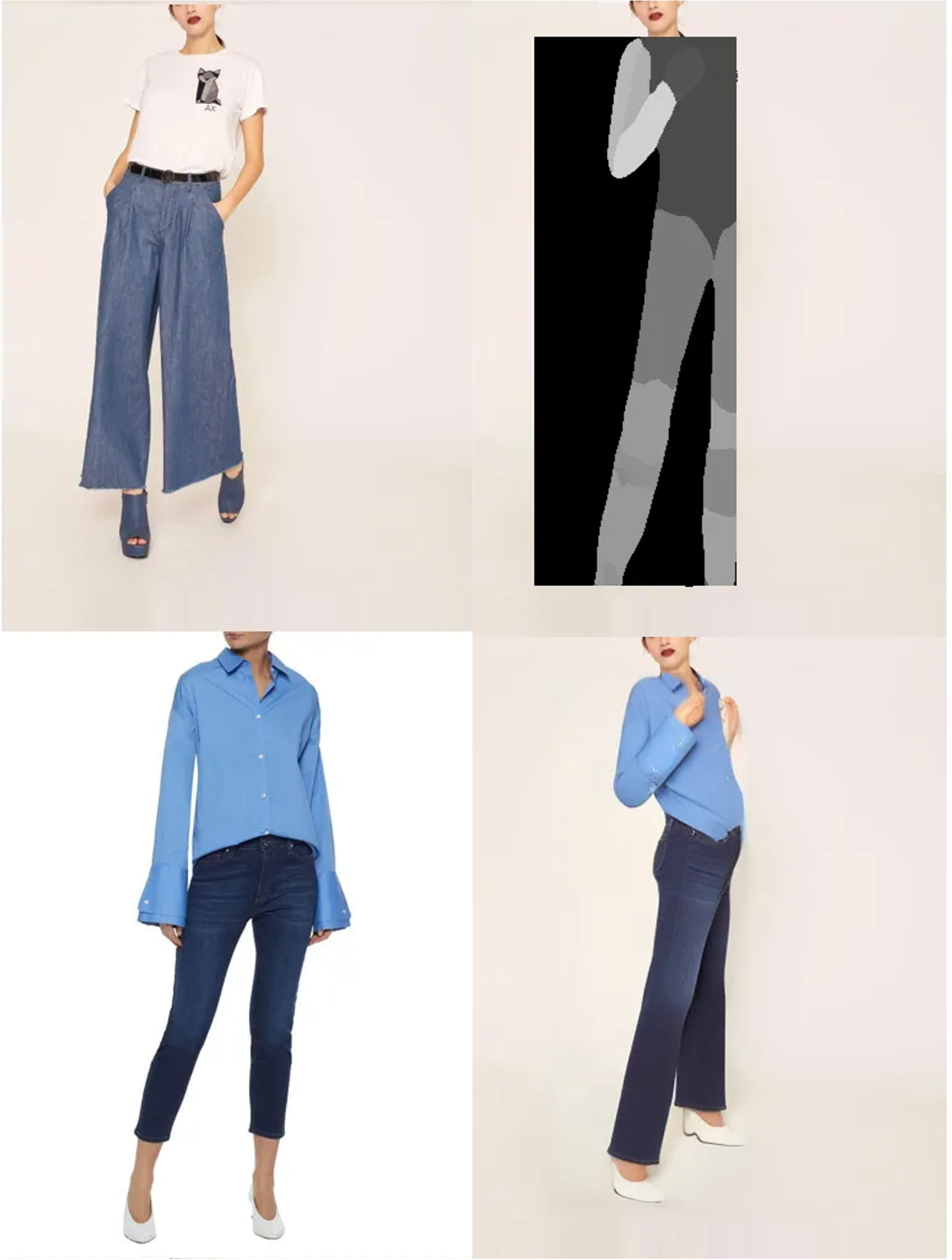}
  \caption{A failure case where the provided pose image is misaligned with the original unmasked context in depth and position.}
  \label{fig:failure}
\end{figure}

\subsection{Qualitative Results}
\subsubsection{Pose Preservation}
\cref{fig:qual pose pre} compares a pose-less baseline model and our model in terms of pose preservation. When parts of the body are masked (especially limb extremities like hands and feet), pose-less model often misinterprets the pose, and in the second example, even alters the T-shirt to a long-sleeve style. In contrast, our model, guided by pose conditions, more accurately preserves the original pose and retains the clothing style.

\subsubsection{Pose Control}
In \cref{fig:qual pose control}, we illustrate how our model adapts to different poses based on the input pose control signals. We compare our results with IDM-VITON ( we use a new pose image as its input during inference to guide generation). Our method effectively generates poses that differ from those in the original source images, offering flexible user-defined poses. While IDM-VITON accurately reflects the target pose, its outputs appear less natural in comparison.

\subsubsection{Ablations}
\cref{fig:ablation} presents the generation results from the five control groups. We observe that pose concatenation introduces unnatural color artifacts, while joint stitching produces reasonable color but tends to overfit within the masked area. The grayscale pose stitching method, in comparison, more accurately captures both the pose and body shape of the person.

Additional comparisons of baseline pose-free/pose-less models and our methods for pose preservation, as well as failure cases of pose-free model are presented in \cref{fig:posecond},

\subsubsection{Limitations}
In \cref{fig:failure}, we illustrate a failure case where the input pose image is misaligned with the context outside the bounding box, causing the generated results to deviate from the expected pose. This highlights the importance of maintaining consistent depth and positioning of the human figure relative to the original pose to achieve more natural results. Extending the model to handle more diverse poses with variations in position and depth will be explored in future work.


\section{Conclusion}
In this work, we explore different pose representations and ways to condition pose as the VTON model input. Experimental results show that stitching grayscale pose map into the masked area of the source image achieves the best generation results and improve the original base model. Meanwhile, it outperforms other baselines that use additional channel to accommodate pose signals. Our method can be flexibly adapted to any VTON models. It is shown to not only preserve the original pose but also generate versatile poses of the original user. 

\clearpage
{
    \small
    \bibliographystyle{unsrt}
    \bibliography{main}
}


\end{document}